\documentclass{ecai} 

\usepackage{latexsym}
\usepackage{amssymb}
\usepackage{amsmath}
\usepackage{amsthm}
\usepackage{booktabs}
\usepackage{enumitem}
\usepackage{graphicx}
\usepackage{color}
\usepackage{listings}
\newtheorem{problem}{Problem}
\usepackage{algorithm}
\usepackage{algpseudocode}
\usepackage{booktabs}
\usepackage{float}
\usepackage{listings}
\usepackage{xcolor}

\usepackage{algorithm}

\algrenewcommand\algorithmiccomment[1]{\hfill{\footnotesize\(\triangleright\) #1}}

\makeatletter
\g@addto@macro\normalsize{%
  \setlength{\abovedisplayskip}{4pt}
  \setlength{\belowdisplayskip}{4pt}
  \setlength{\abovedisplayshortskip}{3pt}
  \setlength{\belowdisplayshortskip}{3pt}
}
\makeatother

\usepackage{enumitem}
\setlist{nosep,leftmargin=*}

\usepackage{microtype}

\newtheorem{theorem}{Theorem}

\newcommand{\BibTeX}{B\kern-.05em{\sc i\kern-.025em b}\kern-.08em\TeX}

\begin{document}


\begin{frontmatter}


\paperid{123} 


\title{Model Recovery at the Edge under Resource Constraints for Physical AI}

\author{\fnms{Bin}~\snm{Xu}\orcid{0000-0000-0000-0000}}
\author{\fnms{Ayan}~\snm{Banerjee}\orcid{0000-0001-6529-1644}}
\author{\fnms{Sandeep K.S.}~\snm{Gupta}\orcid{0000-0002-6108-5584}}

%
\address{IMPACT lab, Arizona State University, Tempe, AZ, USA\\Email: {binxu4, Ayan.Banerjee, Sandeep.Gupta}@asu.edu}


\begin{abstract} 

Model Recovery (MR) enables safe, explainable decision-making in mission-critical autonomous systems (MCAS) by learning governing dynamical equations, but its deployment on edge devices is hindered by the iterative nature of neural ordinary differential equations (NODE), which are inefficient on FPGAs. Memory and energy consumption are the main concern of applying MR on edge devices for real-time running MR. 
We propose MERINDA, a novel FPGA-accelerated MR framework that replaces iterative solvers with a parallelizable neural architecture equivalent to NODEs. 
MERINDA achieves nearly 11× lower DRAM usage and 2.2× faster runtime compared to mobile GPUs. Experiments reveal an inverse relationship between memory and energy at fixed accuracy, highlighting MERINDA’s suitability for resource-constrained, real-time MCAS. 

“The implementation and datasets are publicly available at {\texttt{github.com/ImpactLabASU/ECAI2025}}.”

\end{abstract}

\end{frontmatter}


\section{Introduction}
One of the fundamental advancements in the Artificial Intelligence (AI) revolution is physical AI where computational agents interact with physical systems for the purpose of control and continual learning~\cite{radanliev2021artificial}. Physics-guided predictive inference is central to the future of physical AI, where systems like autonomous vehicles and insulin delivery devices rely on real-time understanding of the underlying physical dynamics. A key technological innovation in physics-guided inference is the ability to extract first-principle–guided dynamical equations from real-world data of physical systems, a process known as model recovery (MR). As opposed to data-driven model learning, the recovered model is a digital twin (DT) that can be used for online monitoring of safety, integrity and unknown errors and feedback from forward simulation of the DT can be used by the system to adjust responses~\cite{tao2018digital,xu2025accelerated}. In this regard, an important requirement of MR techniques arises from the necessary response time of physical AI applications as shown in Table~\ref{tbl:applications}.

Given the data deluge problem in physical AI~\cite{harris2022tesla}, executing MR methods in cloud creates several bottlenecks related to data transfer (Table \ref{tbl:applications} shows that cloud data transfer time in most physical AI applications is close to response time), and energy which hampers response time and operational costs. A potential solution to the data deluge problem is edge AI, where inference is partially offloaded to edge devices to reduce data transfer, storage, and energy costs. However, this introduces higher computation costs, as edge devices (e.g., mobile GPUs) typically have far fewer computational resources compared to cloud GPUs. To the best of our knowledge, MR techniques have not been evaluated for their computational performances in the edge devices. In this paper, we focus on implementation of MR techniques in the edge under memory and energy constraints, characterize their response time performance, and evaluate the feasibility of using MR techniques in real time physical AI applications such as Table \ref{tbl:applications}.

Recently, Koopman theory~\cite{brunton2021modern} has been used to develop sparse regression-based solutions for the MR problem with two broad classes of techniques: a) Sparse regression of nonlinear dynamics (SINDY) and its model predictive control version (SINDY-MPC) that deals with external inputs~\cite{kaiser2018sparse}, and b) a combination of sparse regression with initial state estimation using neural architectures such as physics-informed neural ordinary differential equation (PiNODE)~\cite{PiNode}, physics-informed neural networks with sparse regression (PINN+SR)~\cite{chen2021physics}, and Extraction of underlying Models with Implicit Dynamics (EMILY)~\cite{pmlr-v255-banerjee24a}. Modern MR methods are typically accelerated using GPUs via standard multi-processing pipelines (e.g., PyTorch, TensorFlow). However, as seen in Table \ref{tbl:applications}, state-of-the-art (SOTA) multi-processing fail to meet response times of DT learning for various physical AI applications. This motivates a shift toward deploying MR on hardware acceleration devices such as Field Programmable Gate Arrays (FPGAs), which offer reconfigurable, time and energy-efficient execution. While there has been previous attempts at hardware acceleration of SINDY without forcing inputs~\cite{el2024fpga,ma2018optimizing,FPGATrans}, more generally applicable MR methods that are robust to external inputs, such as EMILY, and noise such as PINN+SR remain largely unexplored on FPGAs—highlighting a key research gap in edge-deployable physical AI.  

\begin{table}[t]
\centering
\scriptsize
\caption{Use of Digital Twins to provide feedback physical AI applications. The table lists application specific response time requirements and DT learning times in SOTA GPU system for EMILY~\cite{BanerjeeECAI24}.}
\begin{tabular}{@{}p{0.5in}@{}p{0.05 in}@{}p{0.25 in}p{0.5 in}@{}p{0.1 in}@{}p{0.3 in}p{0.6 in}@{}p{0.05 in}@{}p{0.3 in}@{}p{0.3 in}}
\toprule
\textbf{Domain}& &\textbf{Hazard} & \textbf{Feedback} & & \textbf{Response Time}&\textbf{DT model}& &\textbf{GPU Time} & \textbf{Transfer time}\\
\midrule
Diabetes \cite{BanerjeeECAI24}& &Hypo-glycemia& Change insulin rate &&900s& Bergman minimal metabolic model&  
 &1412s& 27s\\
Cardiac disease~\cite{Salian24Asilomar,Nabar11GeMREM}& &Ischemia& Alert first responders &&100s& ECG generative model& & 452s& 81s\\
Brain sensing~\cite{BanerjeeECAI24,sadeghi2016toward}& &Attention deficit& Audio-visual cues&&33ms& Resistance capacitance model& & 321s&125s\\
\bottomrule
\end{tabular}
\label{tbl:applications}
\end{table}


In this paper, we address this gap by introducing architectural innovations that enable real-time, resource-aware MR on FPGAs, pushing the physical AI into the Edge AI domain.
Several efforts have accelerated ODENet~\cite{Wtanabe} and NODE architectures~\cite{CaiFPGA}, but these focus on fixed-depth models and lack the flexibility required by frameworks like PINN+SR, PiNODE, or EMILY. As shown in Fig.~\ref{fig:FPGA}, these approaches incorporate large, adaptive NODE layers whose forward passes involve solving high-dimensional ODEs—computationally intensive due to their iterative nature. While some works target standalone ODE solver acceleration~\cite{stamoulias2017high,ebrahimi2017evaluation}, they assume fixed coefficients and are unsuitable for dynamic, data-driven MR tasks like PiNODE, where model parameters vary across inputs.


We leverage the theory of neural flows~\cite{bilovs2021neural} to develop MERINDA (Model Recovery In Dynamic Architecture), an alternative neural structure that is mathematically equivalent to the NODE layers used in EMILY, PiNODE, and PINN+SR while being more amenable to FPGA acceleration (Fig. \ref{fig:FPGA}). Instead of using a conventional NODE layer, MERINDA employs a layer of invertible functions designed through a combination of Gated Recurrent Units (GRU)\cite{salem2021gated} and a dense layer of neurons with nonlinear activation functions, enabling efficient and high-speed computations. The key contributions include:  

\begin{figure}[t]
    \centering \includegraphics[trim={0, 100, 0, 0},width=0.8\columnwidth]{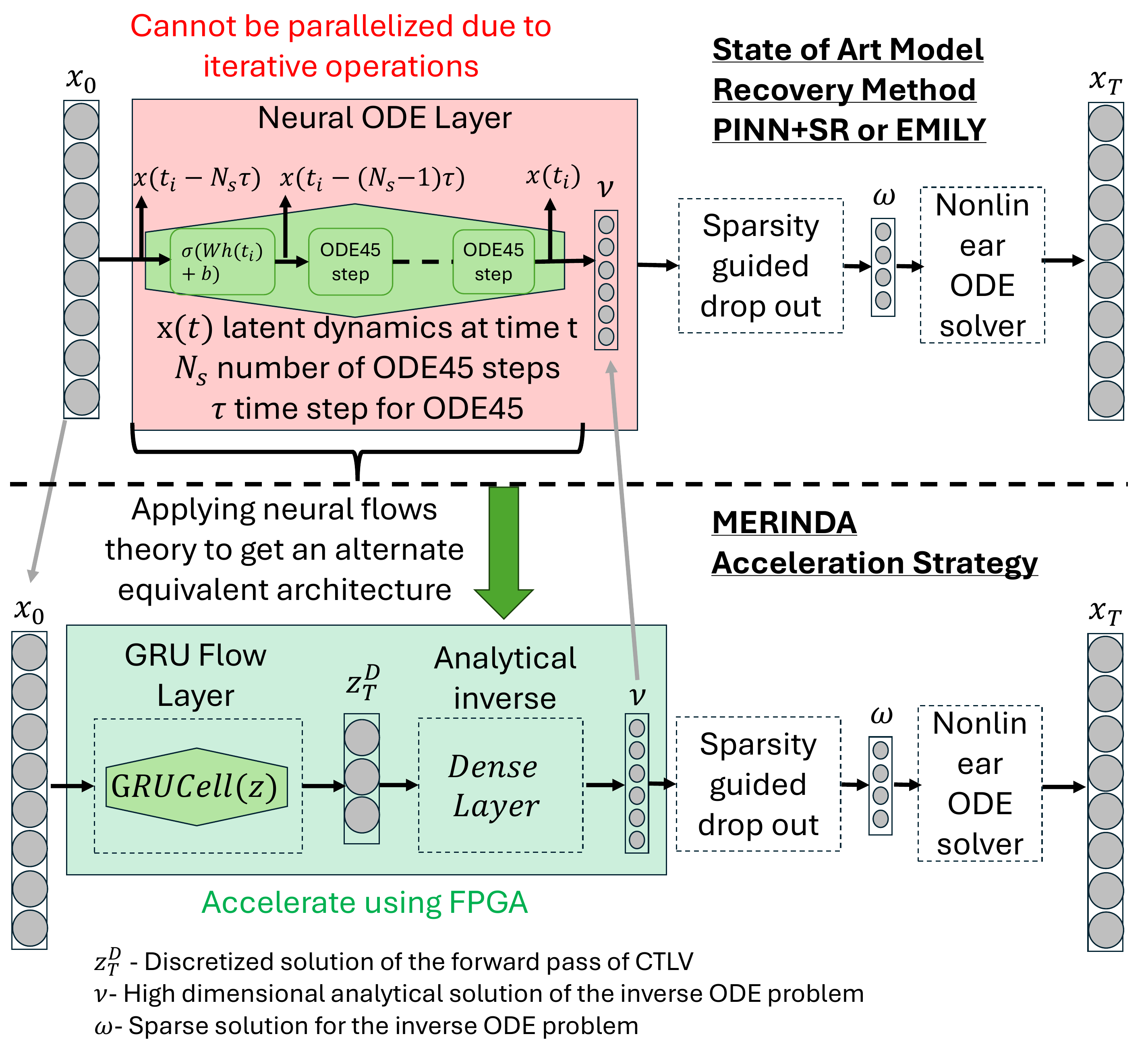}
    \caption{FPGA acceleration strategy using neural flow based equivalent architecture to neural ODEs\cite{chen2018neural}.}
    \label{fig:FPGA} 
\end{figure}

\noindent{\bf 1. } Empirical Benchmarking: We compare MERINDA against state-of-the-art (SOTA) MR techniques, EMILY and PINN+SR, evaluating MR accuracy across four benchmark nonlinear systems, consisting of two real-world and two simulated datasets (see Sec.~\ref{sec:Eval}).

\noindent{\bf 2.} Performance Evaluation in Edge-AI Applications: We assess accuracy, processing speed, energy consumption, and DRAM footprint by comparing MERINDA with SINDY~\cite{zhang2019convergence} and PINN+SR~\cite{LiuPGLoss} in a real-world Automated Insulin Delivery (AID) application (see Sec.~\ref{sec:Eval}).

\noindent{\bf 3. } Comparison with GPU-Based Implementations: We analyze the advantages of MERINDA’s FPGA acceleration by benchmarking it against GPU-based implementations of all three approaches (EMILY, PINN+SR, and SINDY) (see Sec.~\ref{sec:Eval}).

\noindent{\bf 4. } Exploration of the Energy-Memory Tradeoff in MERINDA: We demonstrate that MERINDA exhibits an inherent trade-off between energy consumption and memory usage when maintaining accuracy constraints (see Sec.~\ref{sec:trade}).



\section{Theoretical background}
This section introduces the fundamentals of MR and establishes the approximate equivalence between neural flow architectures and NODEs.

\paragraph{Basics of Model Recovery}
The primary objective of MR is similar to an auto-encoder (Fig. \ref{fig:FPGA}), where given a multivariate time series signal $X(t)$, the aim is to find a latent space representation that can be used to reconstruct an estimation $\Tilde{X}(t)$ with low error. It has the traditional encoder $\phi(t)$ and decoder ($\Psi(t)$) of an auto-encoder architecture\cite{baldi2012autoencoders}. MR represents the measurements $X$ of dimension $n$ and $N$ samples as a set of nonlinear ordinary differential equation model in (\ref{eqn:Model}). 
\begin{equation}
    \label{eqn:Model}
    \scriptsize
    \dot{X} = h(X,U,\theta),
\end{equation}
\vspace{-0.2in}\\
where $h$ is a parameterized nonlinear function that defines the hidden state dynamics, $U$ is the $m$ dimensional external input, and $\theta$ is the $p$ dimensional coefficient set of the nonlinear ODE model.

\noindent{\bf Sparsity:} An $n$-dimensional model with $M^{th}$ order nonlinearity can utilize $\binom{M+n}{n}$ nonlinear terms. A sparse model only includes a few nonlinear terms $p << \binom{M+n}{n}$. Sparsity structure of a model is the set of nonlinear terms used by it~\cite{hastie2015statistical}.

\noindent{\bf Identifiable model:} A model in (\ref{eqn:Model}) is identifiable~\cite{verdiere2019systematic}, if $\exists$ time $t_I > 0$, such that $\forall \theta, \Tilde{\theta} \in \mathcal{R}^p$:
\begin{equation}
    \label{eqn:Ident}
    \scriptsize
    \forall t \in [0,t_I], f(X(t),U(t),\theta) = f(X(t),U(t),\Tilde{\theta}) \implies \theta = \Tilde{\theta}. 
\end{equation}

\noindent (\ref{eqn:Ident}) effectively means that a model is identifiable if two different model coefficients do not result in identical measurements $X$. In simpler terms, this means $\forall \theta_i \in \theta, \frac{dX}{d\theta_i} \neq 0$. We assume that the underlying model is identifiable, where $f$ denotes a general nonlinear function.

\begin{problem}[Sparse Model Recovery]\label{prob:Problem} Given $N$ samples of measurements $X$ and inputs $U$, obtained from a sparse model in (\ref{eqn:Model}) such that $\theta$ is identifiable, recover $\Tilde{\theta}$ such that for $\Tilde{X}$ generated from $f(X,U,\Tilde{\theta})$, we have $||X - \Tilde{X}|| \leq \epsilon$, where $\epsilon$ is the maximum tolerable error.
\end{problem}

\begin{figure*}[h]
    \centering
        \includegraphics[width=0.75\textwidth]{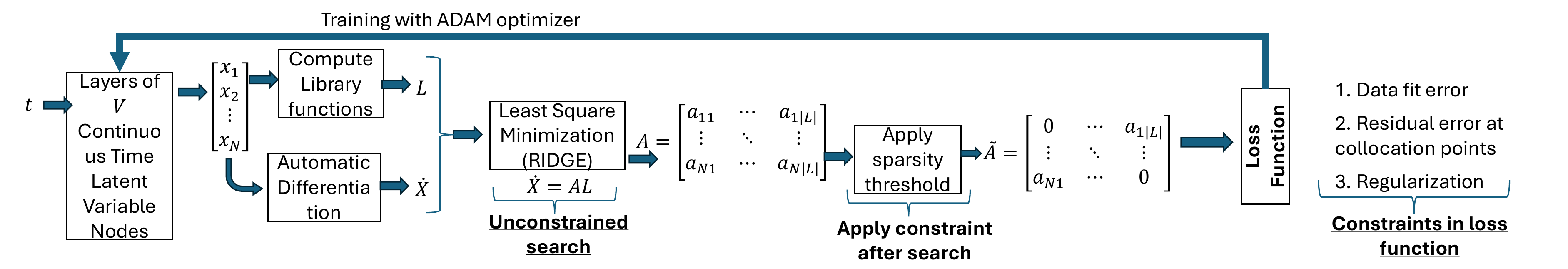}
    \caption{Generic architecture for physics-guided inferencing to be used in physical AI.}
    \label{fig:Arch}
\end{figure*}

\noindent{\bf State-of-the-Art:} Existing MR methods such as PINN+SR and PiNODE (Fig.~\ref{fig:Arch}) express $f$ as a linear combination $\dot{X}=AL$ of nonlinear functions from a library $L$, where $A$ is a coefficient matrix. To build the library, a fully connected continuous-time latent variable (CTLV) layer (e.g., Neural ODE~\cite{PiNode}) converts time $t$ into state estimates $X$. An automatic differentiation (autodiff) layer then estimates derivatives of $X$. Ridge regression identifies matrix $A$~\cite{kaiser2018sparse}, optimizing a loss function that balances data-fit error (real vs. reconstructed data via ODE solver), sparsity (regularization), and residual errors. The main computational costs arise from: i) the CTLV layer's ODE solver, ii) the autodiff RELU layer, and iii) Ridge regression.

\noindent{\bf Role of NODE:} Both EMILY~\cite{pmlr-v255-banerjee24a} and PINN+SR~\cite{chen2021physics} utilize a layer of NODE cells in order to integrate the underlying nonlinear ODE dynamics. NODE is one type of CTLV cell. NODE cell's forward pass is by design the integration of the function $h$ over time horizon $T$ with $N$ samples (Fig. \ref{fig:FPGA}). This effectively requires an ODE solver in each cell of the NODE layer:
\vspace{-0.1in}\\
\begin{equation}
\scriptsize
z(t) = \int\limits^T_0{h(z,u,\theta)dt}, 
\end{equation}
where $z \in Z$ and $u \in U$ are each cells output and input, respectively. 

\paragraph{Neural flows and equivalent architectures to NODE}

According to the theory of neural flows~\cite{bilovs2021neural}, the node layer can be replaced by an approximate solution to $F(t) \approx Z(t)$ in discretized form using recurrent nerual network architectures such as GRU provided that following conditions are satisfied:

\begin{scriptsize}
\begin{equation}
    F(0,u) = Z(0,u), \text{(initial condition), and  }
    F(t,u) \text{  is invertible}.
\end{equation}
\end{scriptsize}

Bilovs et al.~\cite{bilovs2021neural} demonstrated that \( F(t,u) \) can be approximated by replacing the original NODE layer with a GRU layer. However, the GRU layer does not satisfy the invertibility condition, which is essential for ensuring the reconstruction of latent dynamics. To address this, the authors in~\cite{bilovs2021neural} propose the addition of a dense layer, leveraging its property as a universal approximator of nonlinear functions, thereby enabling it to approximate the inverse of \( F(t,u) \).  

MERINDA builds upon and enhances the equivalent architecture proposed in~\cite{bilovs2021neural} by further pruning the dense layer, as illustrated in Fig. \ref{fig:Approach}. The key idea is to optimize the dense layer structure by exploiting the inherent sparsity in the underlying model of the data, thereby improving computational efficiency without compromising accuracy.

\vspace{-0.1 in}
\section{MERINDA Architecture}

\begin{figure}[h]
\centering
\includegraphics[trim={0, 75, 0, 0},width=\columnwidth]{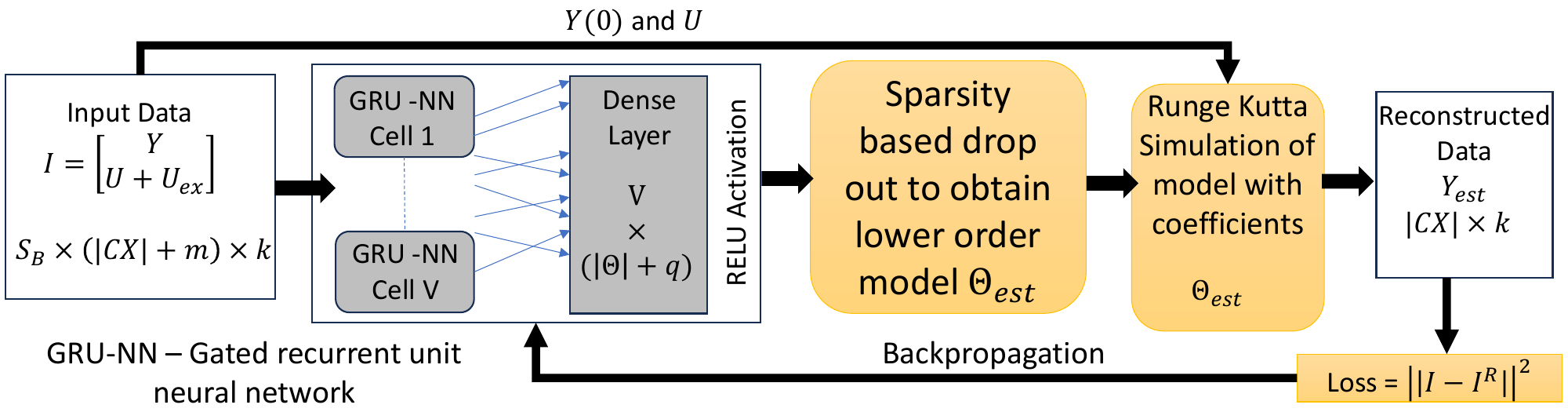} 
\caption{MERINDA: GRU NN-based MR architecture.} \vspace{-0.15in}\
\label{fig:Approach}
\end{figure}

In our approach (Fig.~\ref{fig:Approach}), we extend the GRU architecture to design \textbf{MERINDA}, a neural model tailored for MR. The GRU’s forward pass encodes model coefficients as nonlinear functions of the inputs \( U \) and outputs \( Y \), converting the system dynamics into an overdetermined set of nonlinear equations. To handle redundancy and ensure consistent coefficient estimation, we introduce a dense layer to select a representative subset of equations. This selection is optimized using an \textit{ODE loss}—the mean squared error (MSE) between the observed \( Y \) and the predicted output \( Y_{\text{est}} \), computed by solving the system with an ODE solver: \( \mathbf{SOLVE}(Y(0), \Theta, U) \).

The advanced neural architectures for model recovery in Fig.~\ref{fig:Approach} are implemented by extending the base code available in~\cite{liquid-time-constant-networks}. The training data consists of temporal traces of \( Y \) and \( U \), where \( Y \) is sampled at a rate at least equal to the Nyquist rate for the given application, and \( U \) is sampled at the same rate as \( Y \). The resulting dataset is then divided into batches of size \( S_B \), forming a three-dimensional tensor of size \( S_B \times (|Y|+m) \times k \).  
Each batch is processed through the \textit{GRU-NN} network with \( V \) nodes, generating \( V \) hidden states. A dense layer is then applied to transform these \( V \) hidden states into \( p = |\Theta| \) model coefficient estimates and \( q \) input shift values. The dense layer is structured as a multi-layer perceptron with a ReLU activation function for the model coefficient estimate nodes, whose outputs correspond to the estimated model coefficients. The dense layer maps the \( V \)-dimensional hidden state outputs to \( \binom{M+|X|}{|X|} \), which represents the number of nonlinear terms available for an \( M \)th-order polynomial. A dropout rate of \( |\Theta| \) is applied to ensure that the final number of active output layers corresponds to \( |\Theta| \).  

The estimated model coefficients, along with the initial value \( Y(0) \), are passed through an ODE solver to compute the solution for the nonlinear dynamical equations using the estimated coefficients \( \Theta_{\text{est}} \), initial conditions \( Y(0) \), and inputs \( U \). The solver employs the Runge-Kutta integration method to obtain \( Y_{\text{est}} \). During the backpropagation phase, the network loss is augmented with the ODE loss, defined as the MSE between the original trace \( Y \) and the estimated trace \( Y_{\text{est}} \).

\vspace{-0.1 in}
\subsection{Proof of Equivalence}
We now prove that the architecture in Fig. \ref{fig:Approach} is equivalent to the NODE based MR architecture in Fig. \ref{fig:Arch}.

\noindent\textbf{Neural ODE Formulation.}  
In a NODE network, differentiating the forward pass of a single cell yields the standard ODE form
\begin{scriptsize}
\begin{equation}
\frac{dx(t)}{dt} = h( x(t),u,\theta).
\label{eq:node_diff}
\end{equation}
\end{scriptsize}

\medskip
\noindent\textbf{Koopman Operator Formulation.}  
According to Koopman theory~\cite{sinha2020koopmanoperatormethodsglobal}, the nonlinear dynamics in \eqref{eq:node_diff} can be represented in a potentially infinite-dimensional linear space. This representation involves:  
(i) a Koopman operator $K$,  
(ii) a measurement function $G$, and  
(iii) a set of latent variables $Z$.  
The dynamics are expressed as
\begin{scriptsize}
\begin{equation}
K\,G(t, Z) = \frac{d}{dt} G(t, Z), 
\qquad 
x(t) = G(t, Z).
\label{eq:koopman}
\end{equation}
\end{scriptsize}

\medskip
\noindent\textbf{Mori--Zwanzig (MZ) Formulation.}  
The Koopman representation can be partitioned into two components:  
(a) observable dynamics $G_m$, and  
(b) implicit dynamics $G_i$.  
The coupled system is then written as
\begin{scriptsize}
\begin{equation}
\frac{d}{dt}
\begin{bmatrix}
G_m \\
G_i
\end{bmatrix}
=
\begin{bmatrix}
K_M & K_{IM} \\
K_{MI} & K_I
\end{bmatrix}
\begin{bmatrix}
G_m \\
G_i
\end{bmatrix},
\label{eq:mz_coupled}
\end{equation}
\end{scriptsize}where $K_M$ and $K_I$ represent the self-dynamics of the observable and implicit states, while $K_{IM}$ and $K_{MI}$ capture cross-coupling effects.  

By applying the MZ projection formalism~\cite{hijon2010mori} and Laplace transform, the observable dynamics $G_m$ evolve as
\begin{scriptsize}
\begin{equation}
\frac{dG_m}{dt} 
= K_M G_m 
+ K_{IM} \int_{-\infty}^{t} e^{(t-s)K_I}\, K_{MI}\, G_m(s)\, ds ,
\label{eq:mz_gm}
\end{equation}
\end{scriptsize}where the first term captures the instantaneous effect of observables, while the second term introduces a \emph{memory effect}, expressed as a convolution between the kernel $e^{(t-s)K_I}K_{MI}$ and the past trajectory $G_m(s)$.  
For clarity, stochastic noise terms in the full MZ decomposition are omitted. Once $G_m$ is obtained, the original system state can be approximately reconstructed via a projection operator $P$:
\begin{scriptsize}
\begin{equation}
x(t) \approx P\,G_m(t),
\label{eq:mz_state}
\end{equation}
\end{scriptsize}where $P$ denotes the mapping from observable coordinates back to the physical state space.

\medskip
\noindent\textbf{GRU Formulation.}  
In a GRU cell, the hidden state is defined as
\begin{scriptsize}
\begin{equation}
h(t) = r(t) \circ h(t-1) + \big( 1 - r(t) \big) \circ c(t-1),
\label{eq:gru_update}
\end{equation}
\end{scriptsize}where $r(t)$ is the reset gate, $c(t)$ is the candidate state, and $\circ$ denotes elementwise multiplication.  
Rearranging, we obtain
\begin{scriptsize}
\begin{equation}
h(t) = h(t-1) + \big( 1 - r(t) \big) \circ \big(c(t-1) - h(t-1)\big).
\label{eq:gru_rearranged}
\end{equation}
\end{scriptsize}

\medskip
The recurrence in \eqref{eq:gru_rearranged} has the same structure as the Mori--Zwanzig approximation in \eqref{eq:mz_gm}, where the convolutional memory term models the implicit dynamics. Hence, for every NODE-based architecture, there exists an equivalent GRU-based formulation with parameters that capture the effects of both observable and implicit dynamics.

\subsection{Energy-Memory tradeoff in Physics Guided MR}
\label{sec:trade}
 A key technological innovation towards physical AI is MR, i.e., the extraction of underlying dynamical equations ($f$,$\theta$) from multi-variate data ($X(t)$) generated by real world systems~\cite{kaiser2018sparse}. 
 Acceleration strategies for MR has to carefully navigate three objectives: i) accuracy, the recovered model should reconstruct real data with less error, ii) computational energy, the MR acceleration should be energy efficient, and iii) memory, MR acceleration should have restricted DRAM footprint. Akin to the CAP theorem in distributed computing~\cite{Brewer00CAP}, we introduce the MACE theorem for physical guided learning: 

\begin{theorem}
\label{theo:MACE}
 {MACE theorem:} An acceleration strategy for physics informed model recovery can only guarantee two out of three constraints of memory footprint, recovery accuracy, and computational energy efficiency.   
\end{theorem}

While prior work has explored energy-accuracy and memory-accuracy trade-offs, we identify a fundamental energy-memory trade-off when accuracy is held constant, grounded in Koopman theory. According to the MACE theorem, this trade-off arises from the ability to exchange nonlinearity with dimensionality: higher-order nonlinearities can be represented using more state variables. Increasing nonlinearity increases memory, while increasing dimensionality raises energy consumption—establishing a core trade-off in physics-guided learning. In contrast, data-driven learning exhibits no such structured relationship, with energy and memory scaling jointly.


\subsubsection{MERINDA Memory model} For a $N$ dimensional system, if the size of the CTLV layer is $V$ and if each CTLV node requires $b_c$ bits to store weights, then total memory required for CTLV layer is $NVb_c$ assuming full connectivity. The memory requirement of the autodiff $N$ and library $L$ layer depends on the chosen order of nonlinearity $M$ and is given by $M+N\choose M$ for the library and $N$ for the autodiff layers. The ridge regression requires to store the $N\times {M+N\choose M} $ matrix $A$ and perform its pseudo inverse, which requires further $N \times {M+N\choose M} + max(N^2,M^2)$ memory. Hence the overall memory requirement is given by -
\begin{scriptsize}
\begin{equation}
\sigma_M = N V b_c 
+ \Big( N + {M+N\choose M} + N \times {M+N\choose M} + \max(N^2, M^2) \Big) b_r
\end{equation}
\end{scriptsize}
where $b_r$ is the bits required to store one real number.

\subsubsection{MERINDA Energy Model}
Energy depends on power consumption of each forward and backward pass and the number of epochs used to train the entire architecture. The power consumption of each forward / backward pass of CTLV node $p^f_c, p^b_c$ is higher since it requires solution of ODE while that of the autodiff node $p^f_a, p^b_a$ and RELU node $p^f_l,p^b_l$ are much lower. The actual power values are architecture dependent, GPU or FPGA. If $H$ is the number of samples used in the data, then the Ridge regression requires $H \times {N+M \choose M}^2$ multiplication operations only in the forward pass. Power is also consumed in computing the loss function with the major contributor being the ODE solver. The execution time depends on the sparsity of the matrix $A$ and the stiffness $k$ of the ODE defined by $A$, which is proportional to the inverse of the maximum diagonal term in $A$. For the standard ODE45\cite{senan2007brief} solution one iteration involves solving a difference equation for each dimension, which entails in the worst case $N {M+N \choose M}$ multiplication operations. If the ODE is solved for a horizon of $H$ samples then the total number of operations are $H N {M+N \choose M}$. The total energy consumption depends on the number of epochs $T$ and is given by: 

\begin{scriptsize}
\begin{eqnarray}
& \sigma_E = \big(N V w_c (p^f_c + p^b_c) + N(p^f_a + p^b_a) +  {M+N\choose M}  (p^f_l + p^b_l) \nonumber \\
& \quad + (H \times {N+M \choose M}^2)p_m + H N {M+N \choose M} p_m \big) T
\label{eq:equ6}
\end{eqnarray}
\end{scriptsize}
where $w_c$ is the number of convolution weights involved in the computation per epoch, $p_m$ is the power consumption of a single multiplication operation dependent on platform i.e. GPU or FPGA.

\subsubsection{Proof of MACE theorem}

In most typical applications $N > M$. For high values of $N$, the memory model from (5) reduces to $\mathcal{O}(N^{2})$. Hence, memory is independent of polynomial order for high number of state variables.

On the other hand, for high $N$, the energy model from (6) reduces to $\mathcal{O}(M^{N})$. According to Koopman theory, $M$ can be reduced by increasing $N$. When $N$ increases, $\mathcal{O}(M^{N})$ initially increases, but as $M$ reduces, $\mathcal{O}(M^{N})$ starts decreasing. Hence, as $N$ increases, memory keeps on increasing but energy, after a certain point, starts reducing.

On the other hand, as $N$ reduces, $M$ starts increasing. This has little effect on memory since the overpowering term is still $\mathcal{O}(N^{2})$, which is independent of $M$. However, energy starts increasing exponentially and, for high $M$ when $N$ reduces to $1$, energy increases linearly.
This can be observed in Fig.~4 (right plot).

In most practical applications, $N > M$. For large $N$, the memory model from (5) simplifies to $\mathcal{O}(N^{2})$, indicating that memory cost is independent of the polynomial order and depends only on the number of state variables.  

For the energy model, (6) reduces to $\mathcal{O}(M^{N})$ at high $N$. According to Koopman theory, increasing $N$ allows for a reduction in $M$. Consequently, while $\mathcal{O}(M^{N})$ initially increases with $N$, the reduction in $M$ eventually dominates, causing $\mathcal{O}(M^{N})$ to decrease. Thus, as $N$ increases, memory usage grows quadratically, but energy cost eventually decreases beyond a certain threshold.  

Conversely, when $N$ decreases, $M$ increases. Since the dominant term in the memory model remains $\mathcal{O}(N^{2})$, memory is largely unaffected by changes in $M$. However, the energy cost grows rapidly: for moderate $M$ it increases exponentially, and in the extreme case where $N = 1$, the energy scales linearly with $M$. This theoretical behavior is consistent with the empirical trends shown in Fig.~4 (right plot).

\subsection{Accelerator Design Exploration}
Algorithm~\ref{alg:algo1} presents the pseudo-code of a GRU cell followed by a dense layer, applied iteratively over a multivariate input sequence. 
The output of Algorithm~\ref{alg:algo1} is the predicted sequence $\texttt{y\_pred}[t]$, accompanied by intermediate hidden states and model parameters. The following is a description of the key variables used:

\begin{algorithm}
\caption{Pseudo-code for Iterative GRU Cell and Dense Layer Computation}
\label{alg:algo1}
\footnotesize               
\setlength{\textfloatsep}{6pt} 
\scriptsize 
\begin{algorithmic}[1]
\setlength{\itemsep}{0.2ex} \setlength{\parsep}{0pt} \setlength{\parskip}{0pt} 
\For{each time step $t$ in input sequence}
    \State $x[t] \gets \text{reshape}(X[t, :])$
    \State $\text{concat} \gets \begin{bmatrix} a\_prev[t{-}1] \\ x_t \end{bmatrix}$  \textbf{\Comment{Concat input and hidden state}}

    \State $r[t] \gets \sigma(W_r \cdot \text{concat} + b_r)$ \textbf{\Comment{Reset gate}}
    \State $z[t] \gets \sigma(W_z \cdot \text{concat} + b_z)$ \textbf{\Comment{Update gate}}

    \State $cc[t] \gets \tanh\left(W_a \cdot \begin{bmatrix} r[t] \cdot a[t{-}1] \\ x_t \end{bmatrix} + b_a\right)$ \textbf{\Comment{Candidate cell}}
    
    \State $c[t] \gets z[t] \cdot cc[t] + (1 - z[t]) \cdot c[t{-}1]$ \textbf{\Comment{Cell state}}
    \State $a[t] \gets c[t]$ \textbf{\Comment{Hidden state}}

    \State $y\_pred[t] \gets \text{softmax}(W_y \cdot a[t] + b_y)$ \textbf{\Comment{Dense Layer}}
\EndFor
\end{algorithmic}
\end{algorithm}


\begin{itemize}
    \item \textbf{\texttt{x[t]}}: Input vector at time step $t$.
    \item \textbf{\texttt{a\_prev[t{-}1]}}: Hidden state from the previous step.
    \item \textbf{\texttt{concat[]}}: Concatenation of $x_t$ and $a[t{-}1]$ for gate computation.
    \item \textbf{\texttt{r[t]}}: Reset gate at $t$, controls forgetting of past state.
    \item \textbf{\texttt{z[t]}}: Update gate at $t$, controls retention of past state.
    \item \textbf{\texttt{cc[t]}}: Candidate state at $t$, modulated by the reset gate.
    \item \textbf{\texttt{c[t]}}: Cell state, interpolation of candidate and previous state 
    \item \textbf{\texttt{a[t]}}: Current hidden state, input to dense layer.
    \item \textbf{\texttt{y\_pred[t]}}: Model output at $t$, from softmax layer.
    \item \textbf{\texttt{W\_r}}, \textbf{\texttt{W\_z}}, \textbf{\texttt{W\_a}}, \textbf{\texttt{W\_y}}: Weight matrices.
    \item \textbf{\texttt{b\_r}}, \textbf{\texttt{b\_z}}, \textbf{\texttt{b\_a}}, \textbf{\texttt{b\_y}}: Corresponding bias vectors.
\end{itemize}


Algorithm~\ref{alg:algo2} presents the corresponding high-level synthesis\cite{coussy2009introduction} (HLS) C++ implementation executed on \texttt{AMD Vitis}. Due to the 2D nature of most computations, several intermediate variables are introduced. Below is a summary of the variables used in Algorithm~\ref{alg:algo2}:


\begin{itemize}
    \item \textbf{\textit{H}}: Hidden units in the GRU layer.
    \item \textbf{\textit{I}}: Input vector size (features).
    \item \textbf{\textit{O}}: Output vector size (classes).
    \item \textbf{\texttt{concat[]}}: Concatenation of previous hidden state and current input for gate computation.
    \item \textbf{\texttt{mod\_concat[]}}: Reset-gated hidden state combined with current input for candidate computation.
    \item \textbf{\texttt{sum\_r}}, \textbf{\texttt{sum\_z}}: Accumulators for reset and update gates.
    \item \textbf{\texttt{sum}}: Accumulator for candidate state.
    \item \textbf{\texttt{logits[]}}: Pre-softmax output scores.
\end{itemize}


\subsubsection{Computation Optimization}
\label{sec:gru_loop_dependency_compact}
To enable high-throughput GRU execution, we design a deeply pipelined hardware architecture guided by the loop dependencies. Here L4 depends on L2 and L3, L5 depends on L2 and L4, L7 depends on L5 and L6, L8 depends on L2, L4, and L7, and L9 depends on L8. All optimizations are realized using HLS, which allows us to translate high-level C/C++ descriptions into RTL while automatically handling scheduling and resource allocation. The outer loop over time steps (Algorithm~\ref{alg:algo2}, Line~2) is pipelined with \texttt{\#pragma HLS PIPELINE II=1}, launching a new GRU cell every clock cycle. Internal stages—such as gate computation (Line~10) and cell state update (Line~33)—are decomposed into sub-loops and parallelized using \texttt{\#pragma HLS UNROLL} (L2--L9), enabling concurrent MAC operations across hidden/input dimensions (e.g., Lines~13--14).

To overlap compute stages, we apply \texttt{\#pragma HLS DATAFLOW} before the outer loop, enabling stream-based communication between stages (e.g., L2→L3→L4). This fusion of task-level dataflow and loop-level pipelining achieves initiation interval (II) = 1, maximizing throughput for real-time inference.

\subsubsection{Memory Optimization}

To reduce off-chip memory access and improve data locality, we use HLS pragmas to allocate key variables to on-chip buffers, guided by the loop dependencies in Sec.~\ref{sec:gru_loop_dependency_compact}. The hidden state buffer \texttt{a\_prev}, reused across L2$\rightarrow$L5$\rightarrow$L8, is fully partitioned with \texttt{\#pragma HLS ARRAY\_PARTITION complete} for parallel access.

Single-timestep intermediates (\texttt{concat}, \texttt{mod\_concat}, \texttt{r}, \texttt{z}, \texttt{cc}, \texttt{c}, \texttt{a}) span L2--L8 and are mapped to either registers or BRAM based on reuse patterns. For instance, \texttt{concat} (L2--L4) is stored in registers for fast access in unrolled loops (Lines 13--14), while \texttt{mod\_concat} (L5--L7, Lines 20, 23, 28) is placed in BRAM to exploit spatial locality and reduce routing overhead.

Moreover, \texttt{a} computed in L8 is directly consumed in the dense layer (L9) and softmax function (Line 46), avoiding unnecessary DRAM writes. These memory-aware optimizations, combined with pipeline-level control (\texttt{\#pragma HLS PIPELINE} and \texttt{DATAFLOW}), ensure tightly coupled compute stages with a one-cycle initiation interval (II=1), maximizing throughput in real-time FPGA execution.

\section{Evaluation and Results}  
\label{sec:Eval}
 Our evaluation aims are to:  
 a) Empirically demonstrate that MERINDA is equivalent to EMILY and PINN+SR.  
 b) Assess the benefits of MERINDA by comparing with SOTA MR techniques, Sindy~\cite{kaiser2018sparse} and PINN+SR~\cite{chen2021physics} on both FPGA and GPU platforms.  
 c) Exploring tradeoff between energy consumption and memory.

\begin{algorithm}[t]
\caption{Customized GRU Accelerator with Dense Layer}
\label{alg:algo2}
\footnotesize               
\setlength{\textfloatsep}{6pt} 
\scriptsize 
\begin{algorithmic}[1]
\setlength{\itemsep}{0.2ex} \setlength{\parsep}{0pt} \setlength{\parskip}{0pt} 
\State Before For Loop \Comment{HLS DATAFLOW}
\For{each time step $t$ in input sequence} \Comment{L1, HLS PIPELINE II=1} 

    \State\Comment {\textbf{Concat input and hidden state}}
    \For{$i = 0$ to $H-1$} \Comment{L2, HLS UNROLL}
        \State $\text{concat}[i] \gets a\_prev[i]$
    \EndFor
    \For{$i = 0$ to $I-1$} \Comment{L3, HLS UNROLL}
        \State $\text{concat}[H + i] \gets X[t][i]$
    \EndFor
    
    \State\Comment {\textbf{Reset and Update gates}}
    \For{$i = 0$ to $H-1$} \Comment{L4, HLS UNROLL}
        \State $sum\_r \gets br[i],\quad sum\_z \gets bz[i]$
        \For{$j = 0$ to $H + I - 1$} \Comment{HLS UNROLL}
            \State $sum\_r \mathrel{+}= Wr[i][j] \cdot concat[j]$
            \State $sum\_z \mathrel{+}= Wz[i][j] \cdot concat[j]$
        \EndFor
        \State $r[i] \gets \sigma(sum\_r),\quad z[i] \gets \sigma(sum\_z)$
    \EndFor
    
    \State\Comment {\textbf{Compute candidate cell state}}
    \For{$i = 0$ to $H-1$} \Comment{L5, HLS UNROLL}
        \State $mod\_concat[i] \gets r[i] \cdot a\_prev[i]$
    \EndFor
    \For{$i = 0$ to $I-1$} \Comment{L6, HLS UNROLL}
        \State $mod\_concat[H + i] \gets X[t][i]$
    \EndFor
   
    \For{$i = 0$ to $H-1$} \Comment{L7, HLS UNROLL}
        \State $sum \gets ba[i]$
        \For{$j = 0$ to $H + I - 1$} \Comment{HLS UNROLL}
            \State $sum \mathrel{+}= Wa[i][j] \cdot mod\_concat[j]$
        \EndFor
        \State $cc[i] \gets \tanh(sum)$
    \EndFor

    \State \Comment{\textbf{Update cell state and hidden state}}
    \For{$i = 0$ to $H-1$} \Comment{L8, HLS UNROLL}
        \State $c[i] \gets z[i] \cdot cc[i] + (1 - z[i]) \cdot a\_prev[i]$
        \State $a[i] \gets c[i]$
        \State $a\_prev[i] \gets a[i]$
    \EndFor
   
    \State \Comment{\textbf{Dense Layer}}
    \For{$i = 0$ to $O-1$} \Comment{L9, HLS UNROLL}
        \State $logits[i] \gets by[i]$
        \For{$j = 0$ to $H-1$} \Comment{HLS UNROLL}
            \State $logits[i] \mathrel{+}= Wy[i][j] \cdot a[j]$
        \EndFor
    \EndFor
 
    \State $y\_pred[t][:] \gets \textbf{SoftmaxLayer}(logits)$
\EndFor
\end{algorithmic}
\end{algorithm}

\begin{table*}[thbp]
\centering
\scriptsize
\caption{Performance comparison among FPGA, Mobile GPU, and GPU implementations across various applications.}
\label{tbl:fpga_gpu_comparison_applications}
\resizebox{\textwidth}{!}{%
\begin{tabular}{lccccccccc ccc}
\toprule
\textbf{System} & \multicolumn{3}{c}{\textbf{Runtime (s)}} & \multicolumn{3}{c}{\textbf{Avg Power (W)}} & \multicolumn{3}{c}{\textbf{DRAM Footprint (MB)}} & \multicolumn{3}{c}{\textbf{Energy (J)}}\\
\cmidrule(lr){2-4} \cmidrule(lr){5-7} \cmidrule(lr){8-10} \cmidrule(lr){11-13}
& FPGA & Mobile GPU & GPU & FPGA & Mobile GPU & GPU & FPGA & Mobile GPU & GPU & FPGA & Mobile GPU & GPU \\
\midrule
AID         & 56.63 & 488.32 & 592.26 & 4.74 & 5.85 & 62.00 & 192.36 & 1870.10 & 1695.09 & 268.43 & 2856.67 & 36719.98 \\
Lotka       & 21.23 & 421.16 & 366.86 & 4.88 & 4.99 & 63.13 & 213.00 & 1951.47 & 2041.55 & 103.54 & 2101.57 & 23159.73 \\
Lorenz      & 21.59 & 414.15 & 224.00 & 4.95 & 6.28 & 63.00 & 268.32 & 1879.61 & 2191.50 & 106.78 & 2600.46 & 14112.00 \\
Pathogenic  & 20.74 & 341.21 & 201.44 & 4.91 & 6.48 & 64.00 & 289.18 & 3604.28 & 3960.55 & 101.73 & 2210.89 & 12892.35 \\
F8          & 20.80 & 59.00 & 183.63  & 4.88 & 6.93 & 64.00 & 352.82 & 4146.19 & 4568.01 & 101.46 & 408.69 & 11752.00 \\
\bottomrule
\end{tabular}%
}
\end{table*}
\vspace{-0.1 in}
 \subsection{Case Studies and Data}\label{AA}
 \noindent{\bf Simulation case studies:} Chaotic Lorenz and F8 Cruiser case studies were obtained from~\cite{kaiser2018sparse} and data was generated by implementing the models in Matlab and using ODE45 to solve the ODEs. 

 \noindent{\bf Real world case studies:} Following datasets were used.


 \noindent {\bf Lotka Volterra:} We used yearly lynx and hare pelts data collected from Hudson Bay Company~\cite{kaiser2018sparse}.

 \noindent {\bf Pathogenic attack system:} The data is available in~\cite{kaiser2018sparse}.

 \noindent {\bf Automated Insulin Delivery (AID):} The datasets are patient data obtained from the OhioT1D dataset available in~\cite{marling2020ohiot1dm}. It is 14 time series data of glucose insulin dynamics. Each time series had a duration of 16 hrs 40 mins which amounts to 200 samples of CGM and insulin. 


\subsection{Implementation Details}\label{AA}  
To evaluate the performance of the FPGA, we conduct experiments on both the GPU and FPGA. The GPU serves as the baseline for performance comparison.  

\noindent{\bf GPU Platform:} The experiments were conducted on a system equipped with an Intel Xeon w9-3475X CPU and an NVIDIA RTX 6000 GPU with 48GB of memory. The model was implemented using TensorFlow 2.10 and Keras 2.10. Power consumption was monitored using {\tt nvidia-smi}, while execution time and DRAM footprint were measured using the {\tt time} and {\tt psutil} libraries.  

\noindent\textbf{Mobile GPU Platform:} Our experiments are conducted on the NVIDIA Jetson Orin Nano Developer Kit, which features a 6-core Arm Cortex-A78AE CPU and 8~GB of LPDDR5 memory. The integrated GPU is based on the NVIDIA Ampere architecture, equipped with 1024 CUDA cores and 32 Tensor Cores.

\noindent{\bf FPGA Platform:} For the FPGA implementation, the experiments were performed on a PYNQ-Z2 board featuring a Dual-Core ARM Cortex-A9 processor and a 1.3M-configurable-gate FPGA. The GRU model was developed from scratch, with the forward pass and backpropagation logic implemented in C++ using High-Level Synthesis (HLS) within AMD’s Vitis tool. The accelerator for the forward pass was integrated with the FPGA board using Direct Memory Access (DMA) to interface with the processing system on the PYNQ-Z2. Power consumption was monitored using Vivado's power report, while execution time and DRAM footprint were recorded using the {\tt time} and {\tt psutil} libraries.

\vspace{-0.1 in}
\subsection{Results}\label{AA}
We evaluate MERINDA against SOTA MR methods in terms of accuracy and benchmark its hardware performance across cloud GPU, mobile GPU, and FPGA platforms. Furthermore, our results highlight an inverse relationship between energy consumption and memory usage in MERINDA.




\vspace{-0.1 in}
\subsubsection{Accuracy Benchmark} 
Compared to SOTA MR methods such as EMILY and PINN+SR, Table~\ref{tbl:SMRFM} shows that MERINDA achieves comparable or even lower errors across four benchmark applications. Accuracy is measured using Mean Square Error between the estimated parameters and the ground truth values. 

\noindent{\bf Key takeaway:} From Table \ref{tbl:SMRFM} we see that for all four benchmark applications, MERINDA architecture is successful in recovering the underlying dynamics with comparable accuracy as SOTA methods which use the standard Tensorflow pipeline for neural network training in GPU based systems.

\begin{table}[H]
\centering
\scriptsize
\caption{Comparison between MERINDA and SOTA MR techniques EMILY and PINN+SR using reconstruction MSE. Errors are absolute values; numbers in parentheses indicate standard deviation.}
\label{tbl:SMRFM}
\begin{tabular}{lccc}
\toprule
\textbf{Applications} & \textbf{EMILY} & \textbf{PINN+SR} & \textbf{MERINDA} \\
\midrule
Lotka Volterra      & 0.03 (0.02)     & 0.05 (0.03)     & 0.03 (0.018) \\
Chaotic Lorenz      & 1.7 (0.6)       & 2.11 (1.4)      & 1.68 (0.4) \\
F8 Cruiser          & 4.2 (2.1)       & 6.9 (4.4)       & 5.1 (2.2) \\
Pathogenic Attack   & 14.3 (12.1)     & 21.4 (5.4)      & 15.1 (10.2) \\
\bottomrule
\end{tabular}
\vspace{-0.1 in}
\end{table}

\begin{table*}[thbp]
\scriptsize
\centering
\caption{Performance comparison among FPGA, Mobile GPU, and GPU implementations for SINDY, PINN+SR, and MR modes.}
\label{tbl:fpga_gpu_comparison}
\resizebox{\textwidth}{!}{%
\begin{tabular}{lcccccccccccccccccc}
\toprule
\textbf{Mode} & \multicolumn{3}{c}{\textbf{Average Error}} & \multicolumn{3}{c}{\textbf{Runtime (s)}} & \multicolumn{3}{c}{\textbf{Avg Power (W)}} & \multicolumn{3}{c}{\textbf{DRAM Footprint (MB)}} & \multicolumn{3}{c}{\textbf{Freq (MHz)}} & \multicolumn{3}{c}{\textbf{Perf/Watt (s/W)}} \\
\cmidrule(lr){2-4} \cmidrule(lr){5-7} \cmidrule(lr){8-10} \cmidrule(lr){11-13} \cmidrule(lr){14-16} \cmidrule(lr){17-19}
& FPGA & Mobile GPU & GPU & FPGA & Mobile GPU & GPU & FPGA & Mobile GPU & GPU & FPGA & Mobile GPU & GPU & FPGA & Mobile GPU & GPU & FPGA & Mobile GPU & GPU \\
\midrule
SINDY     & 7.37 & 4.44 & 3.39 & 1150.72 & 371.84 & 396.8 & 4.73 & 5.88 & 62.00 & 107.27 & 98.2 & 135.71 & 144 & 306 & 1410 & 243.18 & 63.25 & 6.40 \\
PINN+SR   & 8.02 & 4.43 & 3.40 & 1407.62 & 761.2 & 481.86 & 4.81 & 5.45 & 64.01 & 336.41 & 1142.75 & 4436.25 & 165 & 306 & 1410 & 292.58 & 139.67 & 7.52 \\
MR        & 4.60 & 3.07 & 2.90 & 253.97 & 562.75 & 423.21 & 4.91 & 5.53 & 72.00 & 214.23 & 2355.13 & 6118.36 & 173 & 306 & 1410 & 51.72 & 101.77 & 5.88 \\
\bottomrule
\end{tabular}%
}
\end{table*}



\subsubsection{Performance Comparison accross various applications}
We evaluate MERINDA on five benchmark applications spanning varying nonlinear dynamics. As shown in Table~\ref{tbl:fpga_gpu_comparison_applications}, FPGAs consistently operate under 5W average power across all applications. In contrast, GPUs consume over 60W, and mobile GPUs average around 6W. This results in up to a 13× reduction in power consumption compared to GPUs, and substantial energy savings overall. Notably, FPGAs also demonstrate dramatic improvements in runtime efficiency, achieving 8× to 17× faster execution than GPUs depending on the application, while using orders of magnitude less DRAM. 

\noindent{\bf Key takeway:} In depth analysis of Table \ref{tbl:fpga_gpu_comparison_applications} shows that as the number of state variables of the nonlinear dynamic model reduces, the execution time and energy reduces. While the DRAM footprint mostly depends on the complexity of the nonlinearity rather than the number of state variables. The AID system only has a single nonlinear term. Lotka model has two nonlinear terms, while the Lorenz system is a chaotic system. The F8 cruiser has three state variables and multiplicative nonlinearities like Lorenz. The pathogenic system has five variables with several nonlinear terms but is a stable non-chaotic system. We observe a negative correlation between energy and DRAM footprint. 

\noindent\textbf{MERINDA exposes a novel trade-off between energy and memory, where energy is dependent on dimensionality (number of state variables) and memory is dependent on the order of nonlinearity.}

\subsubsection{Performance Comparison in practical example}

As shown in Table~\ref{tbl:fpga_gpu_comparison}, the FPGA implementation achieves substantial efficiency gains compared to the GPU baseline. 
For the MR task, the FPGA offers a 1.67$\times$ speedup in runtime over the GPU (253.97~s vs. 423.21~s), despite operating at significantly lower clock frequencies (173~MHz vs. 1410~MHz).
Additionally, it achieves a 11$\times$ reduction in DRAM footprint (214.23~MB vs. 2355.13~MB vs. 6118.36~MB for MR). Unlike GPUs that rely heavily on external DRAM (e.g., GDDR6 or HBM), FPGAs can store frequently accessed data (e.g., weights, hidden states, intermediates) in on-chip BRAM or registers. The memory optimization strategy uses HLS directives such as \texttt{\#pragma HLS ARRAY\_PARTITION} to fully partition arrays and enable parallel access from LUT-based registers or block RAMs, thereby reducing the need for frequent external memory accesses. In terms of performance-per-watt, the FPGA achieves an 8.8$\times$ improvement over the cloud GPU. The mobile GPU is more energy-efficient than the FPGA, however, it still requires 10$\times$ more DRAM for MR computations.
These findings reinforce conclusions from prior studies such as Cong et al.~\cite{cong2018understanding}, which highlight the FPGA's suitability for energy-efficient computing in resource-constrained environments.

Table~\ref{tab:fpga_resource_breakdown} compares FPGA resource utilization for SINDY, PINN+SR, and MR models with varying hidden sizes. SINDY moderately uses DSPs (53.51\%) and BRAM (34.29\%) but has low LUT demand, indicating a memory-bound design. PINN+SR is logic-intensive (56.27\% LUT) with minimal DSP (18.18\%) and BRAM usage, because it relies heavily on symbolic regression. MR exhibits the highest scalability, with DSP, BRAM, and LUT usage increasing significantly as hidden size grows—reaching up to 72.73\% DSP and 54.16\% LUT for MR(128)—demonstrating its compute-heavy and pipelined architecture. Pipelining design enables the MR model to exploit maximal parallelism, significantly accelerating computation.

\begin{table}[thbp]
\centering
\scriptsize
\caption{Resource utilization of SINDY, PINN+SR, and MR designs with varying hidden sizes on FPGA.}
\label{tab:fpga_resource_breakdown}
\begin{tabular}{lcccc}
\toprule
\textbf{Resource} & \textbf{DSP} & \textbf{BRAM} & \textbf{LUT} & \textbf{FF} \\
\midrule
Available   & 220    & 5040   & 53200   & 106400 \\
SINDY       & 53.51\% & 34.29\% & 17.95\% & 34.00\% \\
PINN+SR     & 18.18\% & 9.63\%  & 56.27\% & 29.47\% \\
MR(16)      & 36.36\% & 11.43\% & 11.75\% & 9.35\% \\
MR(32)      & 72.73\% & 28.57\% & 21.94\% & 13.52\% \\
MR(64)      & 72.73\% & 28.57\% & 37.90\% & 21.18\% \\
MR(128)     & 72.73\% & 34.29\% & 54.16\% & 30.75\% \\
\bottomrule
\end{tabular}
\vspace{-0.1 in}
\end{table}

\begin{figure}[t] 
    \centering
    \scriptsize
    \includegraphics[width=\columnwidth]{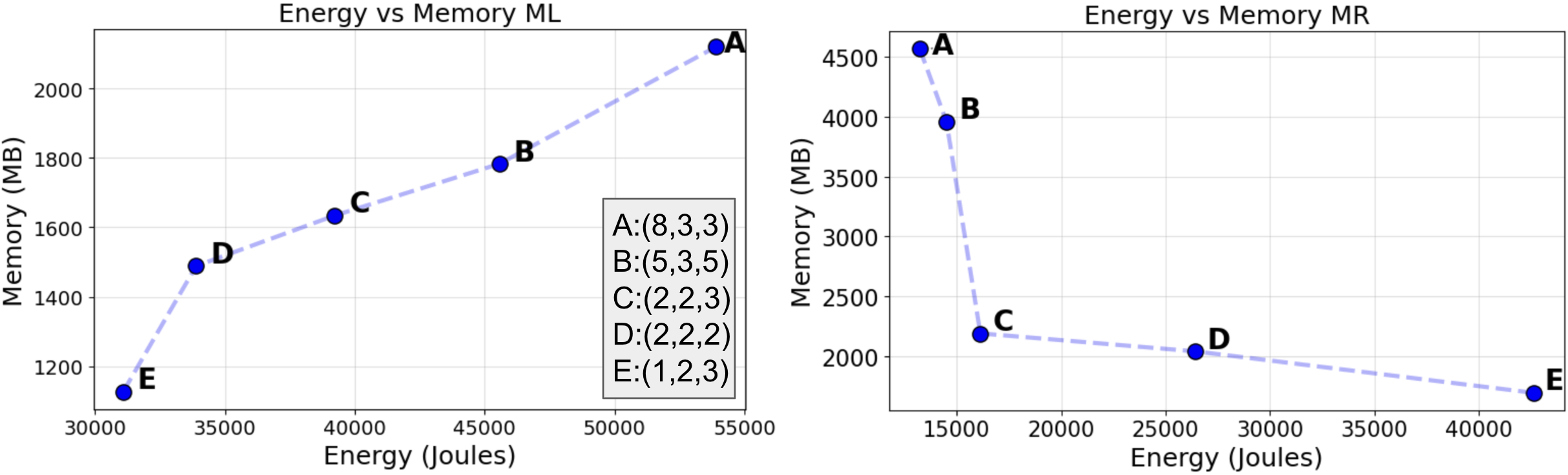}
    \vspace{-0.1 in}
   \caption{Trade-off between memory and energy. The left plot corresponds to SINDY systems, while the right plot corresponds to MR systems. Each dot is annotated with its configuration in the format: (Nonlinear Terms (NL), Polynomial Order (PO), and State Variables (SV)). System complexity details are provided in Table~\ref{tbl:complexity}.}
    \label{fig:memory_enenrgy_tradeoff_comparision}
\end{figure}

\subsubsection{Energy-Memory tradeoff analysis}

From Table~\ref{tbl:fpga_gpu_comparison_applications}, we observe an inverse relationship between energy and memory, further illustrated in Fig.~\ref{fig:memory_enenrgy_tradeoff_comparision}. Our objective is to optimize both metrics by varying the dimensionality $N$ and the degree of nonlinearity $M$, while preserving model recovery accuracy. According to Koopman theory~\cite{sinha2020koopmanoperatormethodsglobal}, reducing the order of nonlinearity ($M$) can be achieved by introducing additional state variables ($N$) that capture differentials of the original states. Thus, maintaining accuracy in model recovery inherently induces an inverse relation between $N$ and $M$.  

In Fig.~4, memory usage remains fixed by the standard FPGA compilation pipeline. To assess energy consumption, we conducted statistical evaluations and report standard deviations (STD) in Table~\ref{tab:ml_mr_std}. Pearson correlation analysis reveals that, for ML, energy and memory are positively correlated with coefficient $R = 0.71$ ($p$-value $=0.04$), whereas for MR, they exhibit a strong negative correlation with $R = -0.83$ ($p$-value $=0.031$). Since both $p$-values are below $0.05$, these correlations are statistically significant.  

Fig.~\ref{fig:memory_enenrgy_tradeoff_comparision} highlights this nontrivial trade-off between energy consumption and memory footprint as system complexity varies across different representations of dynamics. In data-driven ML systems (up plot), increasing the number of nonlinear terms (e.g., $M=8$) leads to a monotonic increase in both energy and memory usage, reflecting the intuitive coupling between model complexity and hardware cost. By contrast, MR systems (down plot) demonstrate a non-monotonic and inverse trend: reducing nonlinearity ($M=1$ or $M=2$) while increasing state variables ($N=2$ or $N=3$) yields substantial energy savings at the expense of higher memory usage.

\noindent{\bf Key takeaway:} By leveraging the inverse relationship between system dimensionality and nonlinearity, one can optimize for energy or memory on hardware platforms without compromising model recovery accuracy.

\begin{table}[ht]
\centering
\scriptsize
\caption{Standard deviation of energy consumption for ML and MR plots across applications.}
\label{tab:ml_mr_std}
\begin{tabular}{lcccccc}
\toprule
\textbf{System} & \textbf{NL} & \textbf{PO} & \textbf{SV} & \textbf{ML Std (J)} & \textbf{MR Std (J)} \\
\midrule
AID         & 1 & 2 & 3 & 160.9 & 219.4 \\
Lotka       & 2 & 2 & 2 & 371.7 & 298.6 \\
Lorenz      & 2 & 2 & 3 & 193.8 & 192.6 \\
Pathogenic  & 5 & 3 & 5 & 317.4 & 137.3 \\
F8          & 8 & 3 & 3 & 348.8 & 135.7 \\
\bottomrule
\end{tabular}
\end{table}

\begin{table}[thbp]
\centering
\scriptsize
\setlength{\tabcolsep}{4pt}
\renewcommand{\arraystretch}{0.95}
\caption{Complexity of systems based on polynomial terms, order, and number of state variables.}
\label{tbl:complexity}
\begin{tabular}{lccc}
\toprule
\textbf{System} & \textbf{\# Poly. Terms} & \textbf{Poly. Order} & \textbf{\# States} \\
\midrule
AID                     & 1 & 2 & 3 \\
Chaotic Lorenz          & 2 & 2 & 3 \\
Lotka-Volterra          & 2 & 2 & 2 \\
Pathogenic Attack Model & 5 & 3 & 5 \\
F8 Cruiser              & 8 & 3 & 3 \\
\bottomrule
\end{tabular}
\end{table}

\vspace{-0.4cm}

\section{Impact Statement}
One of the primary bottlenecks of physical AI is the real-time integration of sensor fusion and perception with foundation model capabilities such as advanced query response. Advanced neural architectures such as transformers for discrete time modeling and liquid foundational models with continuous time modeling capabilities are compute resource hungry and are traditionally slow in embedded hardware. Koopman theory and its associated MZ formulation gives a pathway towards efficient representation of nonlinear dynamics, which is applicable in general to any deep network. Recently a line of work, pioneered by~\cite{BanerjeeECAI24} and EMILY~\cite{pmlr-v255-banerjee24a}, has shown feasibility of using such formulation in model recovery. MERINDA utilizes this breakthrough to enable further acceleration and provides an efficient solution to a major bottleneck for physical AI.

\section{Conclusions}
In this paper, we explored the dynamic reconfiguration of inference architectures on FPGAs, focusing on model recovery. The paper introduced a novel FPGA-based accelerator, MERINDA, designed for physics informed learning. Simulative case study showed that MERINDA is equivalent to PiNODE architecture. While real world study showed significant execution time speedup as compared to GPU. Our evaluations showed the benefits and trade-offs of this approach in terms of power consumption, training time, DRAM footprint, and model accuracy. The results indicate that opportunistic reconfiguration of inference architectures can lead to optimized performance, particularly in mission-critical autonomous systems where both real-time constraints and resource limitations are paramount.

\section*{Acknowledgements}
This work was partially funded by DARPA (AMP, N6600120C4020; FIRE, P000050426), the NSF (FDT-Biotech, 2436801), and the Helmsley Charitable Trust (2-SRA-2017-503-M-B).


\end{document}